\documentclass{llncs}
\usepackage{amsmath,graphicx}
\usepackage[utf8]{inputenc}

\begin{document}
%\ninept
%
\title{Dealing with Topological Information within a Fully Convolutional Neural Network}

\author{Etienne Decencière\inst{1} \and Santiago Velasco-Forero\inst{1} \and Fu Min\inst{2} \and Juanjuan Chen\inst{2} \\ Hélène Burdin\inst{3} \and Gervais Gauthier\inst{3} \and Bruno Laÿ\inst{3} \\ Thomas Bornschloegl\inst{4} \and Thérèse Baldeweck\inst{4}}

\institute{
  MINES ParisTech, PSL Research University, Centre for Mathematical Morphology, France\\
\and L’Oréal Research and Innovation, 550 Jinyu Road, Pudong, Shanghai, China\\
\and ADCIS SA, 3 rue Martin Luther King, 14280 Saint-Contest, France\\
\and L’Oréal Research and Innovation, 1 avenue Eugène Schueller, 93601 Aulnay-sous-Bois, France
}

\maketitle

\begin{abstract}
A fully convolutional neural network has a receptive field of limited size and therefore cannot exploit global information, such as topological information. A solution is proposed in this paper to solve this problem, based on pre-processing with a geodesic operator. It is applied to the segmentation of histological images of pigmented reconstructed epidermis acquired via Whole Slide Imaging.

%%   The emergence of Whole Slide Imaging has brought a revolution to histology. The large image datasets which are produced thanks to this technology require automated analysis tools for their exploitation. Concerning segmentation tasks, deep learning approaches have contributed a significant improvement to the state-of-the-art, as in other fields. However, neural architectures for image segmentation, based on convolutional layers, can only use local information.

%%   A study is presented here on the segmentation of images of pigmented reconstructed epidermis samples used to evaluate and identify the de-pigmenting or pro-pigmenting efficiency of cosmetic ingredients. The goal of the segmentation is to identify two regions corresponding to two specific skin layers: the stratum corneum (SC) and the living epidermis . The boundary between the SC and the background is usually rather difficult to determine, mainly because it can be composed of different layers separated by gaps due to the desquamation process that happens in the SC. Such detached layers are only considered as part of the SC if they constitute an unbroken boundary between the background and the sample. This feature is highly non local, and as such a convolutional neural network cannot enforce it.

%% A method based on mathematical morphology is hereby introduced to solve this problem. Based on geodesic operators, it is evaluated on a database containing 120 images.
\end{abstract}

\section{Introduction}

Image processing and analysis has been revolutionized by the rise of deep learning. For semantic segmentation, deep learning approaches use mainly convolutional neural networks (CNN) \cite{fukushima_neural_1979,lecun_backpropagation_1989}. In the biomedical field, U-Net \cite{ronneberger_u-net:_2015} has become the state-of-the-art method for this task, but other solutions exist, such as SegNet \cite{badrinarayanan_segnet:_2015}. These networks are fully convolutional. Their receptive fields are of limited size. Therefore, they cannot intrinsically process global information, such as topological information \cite{rosenfeld_digital_1979}. We recall that, unlike in networks containing fully connected layers, where the value of each unit depends on the entire input to the network, a unit in a convolutional networks only depends on a region of the input. This region in the input is the receptive field for that unit.

In the following, we present a practical real-world situation where the segmentation result depends on topological information. We show that a classical fully convolutional CNN does not give satisfactory results and propose a solution to this problem.

During the past years, the way of working in the histological field has changed due to the emergence of Whole Slide Imaging solutions that are now available and useful for pathologists but also for dermatologists and cosmetologists. These automated scanners improve digital histology by allowing several hundred slides a day to being acquired through imaging process, and stored, used for second opinions or for automated analysis. This well-established technology provides large sets of image data that need to be automatically processed whatever the amount of generated data. Image analysis methods based on Deep Learning have proved to be extremely useful in this field \cite{bejnordi_diagnostic_2017,wang_deep_2016}.

To circumvent this, we propose a method that allows neural networks based on convolutional layers to take into account non-local information. It is based on a geodesic operator, a popular tool from mathematical morphology. We apply it to the segmentation of whole slide images of pigmented reconstructed epidermis, as described hereafter.

\section{Material}

The images that were collected include pigmented reconstructed epidermis samples used to evaluate and identify the de-pigmenting or pro-pigmenting efficiency of cosmetic ingredients. They have been colored using a Fontana Masson staining, a silver stain that is used to highlight melanin and to also reveal skin layer morphology. Their sizes are diverse and can reach up to 20 million pixels. The ground truth (GT) has been obtained by an automatic method developed by the ADCIS company, whose results were manually edited and modified by L’Oréal experts, when needed. On those images, the goal of the segmentation is to identify two regions corresponding to two specific skin layers: the \textit{stratum corneum} (SC) and the living epidermis.

The boundary between the SC and the background is usually rather difficult to determine, mainly because it can be composed of different layers separated by gaps due to the desquamation process that happens in the SC. Such layers are only considered as part of the SC if they constitute an unbroken boundary between the background and the sample. This feature is highly non local, and as such a convolutional neural network cannot enforce it.

The resulting database contains 120 color images, coming from 46 different slides. The original images are of variable size, and can contain more than 20 million pixels. The test database has been built with approximately 20\% of the images. The remaining images were used for learning and validation. We took care to transfer all the images from a given slide into the same database.

\begin{figure}[t!]
\begin{minipage}[b]{1.0\linewidth}
  \centering
\centerline{\includegraphics[width=10.1cm]{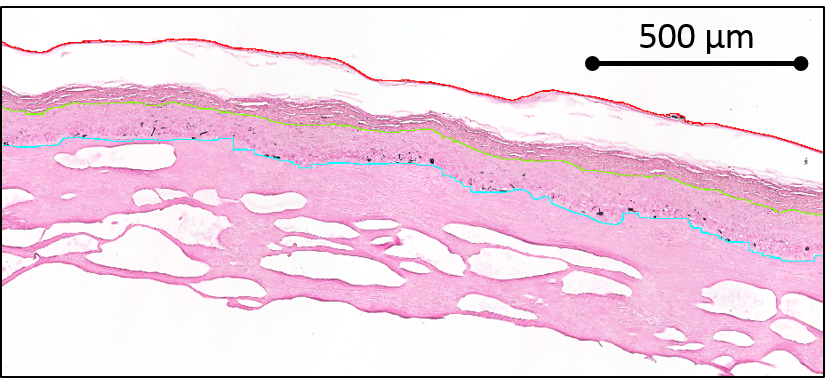}}
%  \centerline{(a) Image example with unbroken}\medskip
\end{minipage}
\begin{minipage}[b]{1.0\linewidth}
  \centering
\centerline{\includegraphics[width=10.1cm]{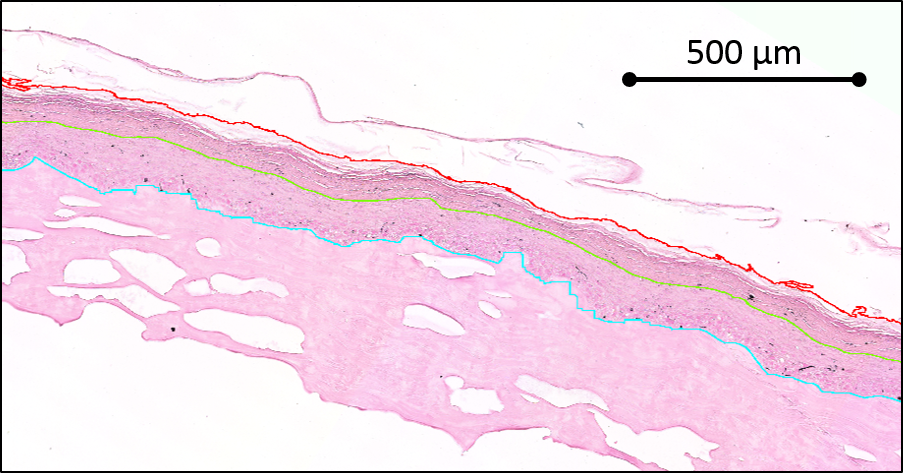}}
   %% \centerline{(a) Result 1}\medskip
\end{minipage}
\caption{Examples of original images with overlayed contours of the reference segmentation (best viewed in colour). Red/top contour: frontier between background and SC. Note that its position can be completely shifted if the detached layer is unbroken (top) or broken (bottom). Green/middle contour: frontier between SC and living epidermis. Cyan/bottom contour: frontier between living epidermis and collagen scaffold (here considered as background).}
\label{fig:examples}
\end{figure}

Given the layered structure of the images, segmenting them into four regions is equivalent to finding three frontiers (see Fig.~\ref{fig:examples}). The top one, between background and SC, has a variable appearance. In some images it corresponds to a contrasted contour; in other to a soft irregular contour. More importantly, its exact position depends on non-local information. Indeed, when a layer of the SC is separated from the rest of the tissue, it will belong to the SC region only if it is connected to the rest of the SC, or if it constitutes an unbroken frontier, going from the left side of the image to the right. Therefore it is not possible to make that decision based uniquely on local information. This kind of situation is illustrated by Fig. \ref{fig:examples}.

The second frontier separates SC from living epidermis. On our images, the distinction between those regions raises from different textures. The third frontier corresponds to the limit between living epidermis and collagen scaffold. Note that the fourth region to be segmented is made up of two compartments: the collagen scaffold that supports the reconstructed skin, and background. The collagen scaffold contains some large “holes” that can locally look as the “holes” within the SC.

The ground truth was generated using an automatic method developed by the ADCIS company, whose results were manually edited and modified by L’Oréal experts, when needed. Given that the top and bottom regions of the ground truth contained both a large white region, which could not be locally differentiated, we decided to only consider three labels: label 1 corresponds to the background (both at the top and the bottom of the images) and collagen scaffold; label 2 to the SC; label 3 to the living epidermis.

\section{Methods}

It was decided to use convolutional neural networks to tackle this problem. During the learning phase, we worked with crops of a given size. After running some tests, the final size of the crops was $512 \times 512$.

The ground truth segmentations, as they contained three labels, were classically represented as an image with three binary channels. Given that background white regions covered the majority of the images, for training we only used the crops that contained at least label 2 or label 3. This procedure is illustrated in Fig.~\ref{fig:comparison}. The resulting set of crops contains 1458 elements. 80\% are randomly picked for learning; the other 20\% are used for validation.

Concerning the loss functions, based on our experience with segmentation using convolutional neural networks, we chose the following loss function between two same length vectors $X$ and $Y$, containing values included between $0$ and $1$:

\[
J_2 = 1 - \frac{XY +  \epsilon}{ X^2 + Y^2 -XY + \epsilon}
\]

where $\epsilon$ is a small constant, used for numerical reasons. This loss is based on the Jaccard index, also called ``intersection over union'' similarity measure, often used to evaluate the quality of a segmentation.

\subsection{Neural network architecture and first results}

%% Several convolutional neural network architectures have been tested, including:
%% \begin{itemize}
%% \item A Pang network \cite{pang_cell_2010};
%% \item U-Net (using a sigmoid in the last activation layer) \cite{ronneberger_u-net:_2015};
%% \item U-Net with dilated convolutions \cite{yu_multi-scale_2015}.
%%   \end{itemize}

After testing several network architectures, the one that gave the best results was U-Net \cite{ronneberger_u-net:_2015} (using a sigmoid in the last activation layer, and using zero-padding in the convolutional layers). Full details are given in section~\ref{ssec:hyper}. The validation loss of the resulting model was $0.085$.

The networks output contains 3 channels. Each one can be interpreted as the probability of a given pixel of belonging to each region. In order to obtain a segmentation, we naturally gave to each pixel the label corresponding to the channel with the highest probability.

\begin{figure}[t!]
\begin{minipage}[b]{1.0\linewidth}
  \centering
\centerline{\includegraphics[width=10.1cm]{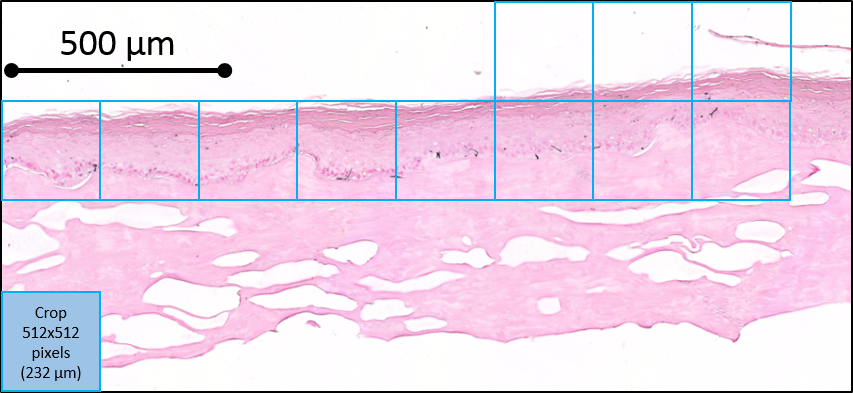}}
%  \centerline{(a) Image example with unbroken}\medskip
\end{minipage}
\begin{minipage}[b]{1.0\linewidth}
  \centering
\centerline{\includegraphics[width=10.1cm]{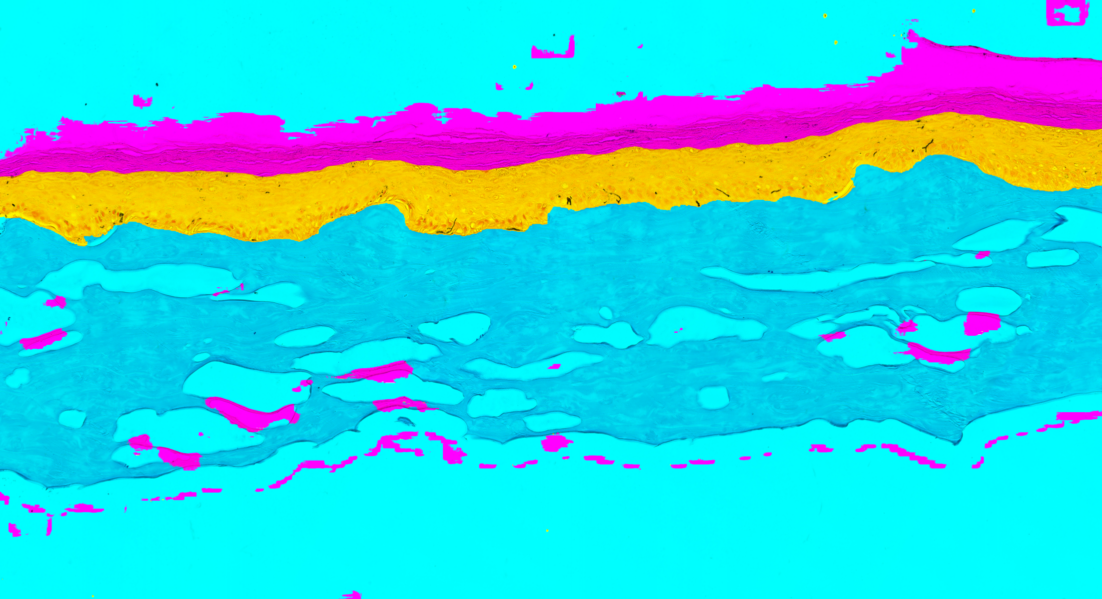}}
   %% \centerline{(a) Result 1}\medskip
\end{minipage}
\begin{minipage}[b]{1.0\linewidth}
  \centering
\centerline{\includegraphics[width=10.1cm]{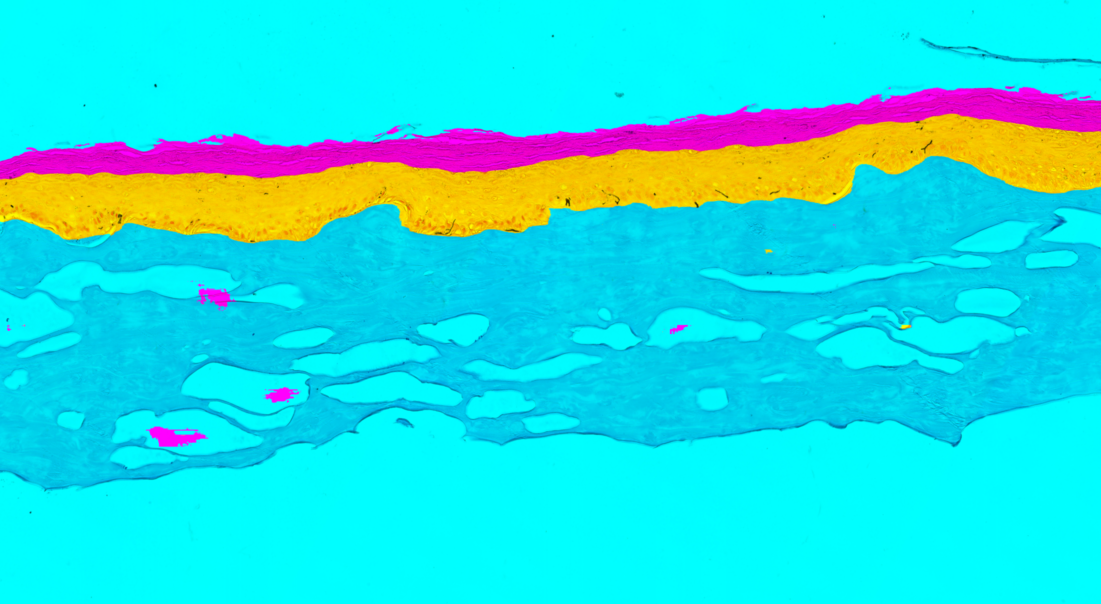}}
   %% \centerline{(a) Result 1}\medskip
\end{minipage}
\caption{Top: original image, showing the selected crops. Middle: results without using global information. Bottom: result using global information, thanks to the presented method. These segmentation results have not been postprocessed. They are overlayed on the original data using the following colour code: SC (magenta), living epidermis (orange) and other regions (cyan).}
\label{fig:comparison}
\end{figure}

A qualitative analysis of the first results showed that the resulting frontiers between SC and living epidermis, on the one hand, and between living epidermis and collagen scaffold, on the other hand, were very satisfactory. However, the frontier between background and SC, in some cases, contained errors. In Fig.~\ref{fig:comparison} (middle) we see that this frontier is incorrectly detected and that the gap between the detached SC layer, on the right, is incorrectly considered as belonging to the SC. These errors span from the fact, as we previously said, that the definition of this frontier is based on non local information.

\subsection{Taking into account non local information in a convolutional neural network}

A new method is proposed here to cope with non local information within convolutional neural networks. It is based on a geodesic reconstruction of the input image from the top and bottom of the image, channel-wise.

%% It is based on the computation of a fourth input channel, containing the connected component of the background which touches the top of the image. This supplementary information allows the network to determine if a white pixel, which seems to be inside the SC, is connected or not to the exterior of the sample. We consider that a pixel belongs to the background if its three values (r, g, b) are larger than a threshold value, here equal to 220. A side effect of this method is that most of the bottom white background is considered as SC. This can be easily corrected with post-processing.

We recall that the geodesic reconstruction \cite{vincent_morphological_1993} of a grey level image $I$ from an image $J_0$ (often called marker) is obtained by iterating the following operation until idempotence:
\[
J_{n+1} = \delta(J_n) \bigwedge I
\]

where $\delta$ is a morphological dilation, here with the cross structuring element, corresponding to the 4-connectivity. Our marker image $J_0$ is equal to $I$ on the first and last rows of the image domain, and to zero elsewhere. The process will ``fill'' the holes within the tissue, and preserve the intensity of the pixels that belong to the connected components of the background that touch the top and bottom of the image. This geodesic operation thus allows to bring topological information, which is essentially global, to a local scope.

In order to recover some of the bright details of the tissue sample, the result of this reconstruction is combined with the initial image by computing their mean. If we call $J$ the result of the above reconstruction, the output image is simply:

\[
F = ( J + I ) / 2 \text{.}
\]

\begin{figure}[t!]
\begin{minipage}[b]{1.0\linewidth}
  \centering
\centerline{\includegraphics[width=10.1cm]{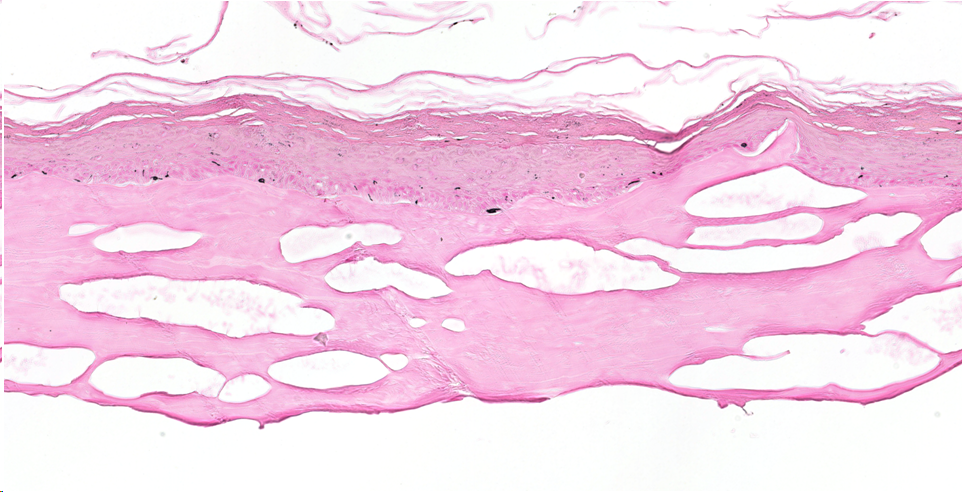}}
\end{minipage}
\begin{minipage}[b]{1.0\linewidth}
  \centering
\centerline{\includegraphics[width=10.1cm]{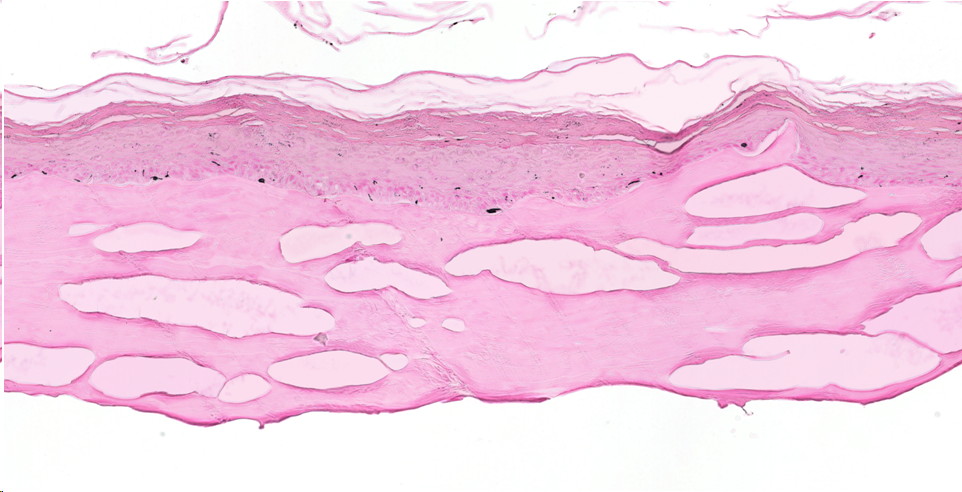}}
   %% \centerline{(a) Result 1}\medskip
\end{minipage}
\caption{Top: original image. Bottom: image after pre-processing based on the geodesic reconstruction. Differences are mainly visible on the holes within the tissue sample.}
\label{fig:example_rec}
\end{figure}

This operator is illustrated in Fig.~\ref{fig:example_rec}. All images follow the same pre-processing (before computing the crops). Learning is done as before, with the same parameters.

The new CNN suppresses the segmentation errors due to the lack of global information on the background / SC boundary. Fig.~\ref{fig:comparison} clearly shows this improvement.

The validation loss of the model is $0.028$, to be compared with the previous value of $0.085$.

%% \subsection{Data augmentation}
%% We tested standard data augmentation methods during the learning process.
%% Using the following transformations:
%% translation (10\%),
%% rotation (5 degrees) and
%% horizontal flip,
%% resulted in a performance decrease. We believe that this is due to the fact that translations and rotations often introduce artifacts in the crops. To check this hypothesis, we tested also with only the horizontal flip in the data augmentation. In this case, the performance decrease disappears, but no significant increase was observed.
%% We concluded that the database is large enough for learning the s

\subsection{Hyper-parameters optimization and data augmentation}
\label{ssec:hyper}

We tuned the hyper-parameters of our system through manual grid search using the validation dataset. The parameters of the final model are: Optimizer: adadelta \cite{zeiler_adadelta:_2012}, with default parameters (learning rate: 1; rho: 0.95; epsilon: 10-8; decay: 0); epochs: 200; patience: 20, batch size: 4. The initial number of convolutional filters of the U-Net network is $16$ (instead of $64$ in the original paper), resulting in a network with  1,962,659 parameters.

We also tested several standard data augmentation methods, but they did not bring any improvement. We believe that this result means that our database constitutes a representative sampling of our image domain.

\subsection{Post-processing}

The current results are already satisfactory. There are however a few defects in the resulting segmentation (as can be seen in Fig.~\ref{fig:comparison}), most of which can be corrected with the following post-processing method:

\begin{enumerate}
\item For the SC and the living epidermis, keep only the largest connected component; for the background, keep the connected components that touch the top and the bottom of the image.
\item Pixels without label are given the label of the closest labelled pixel.
  \end{enumerate}

\section{Results}

\begin{figure}[t!]
\begin{minipage}[b]{0.48\linewidth}
  \centering
\centerline{\includegraphics[width=5cm]{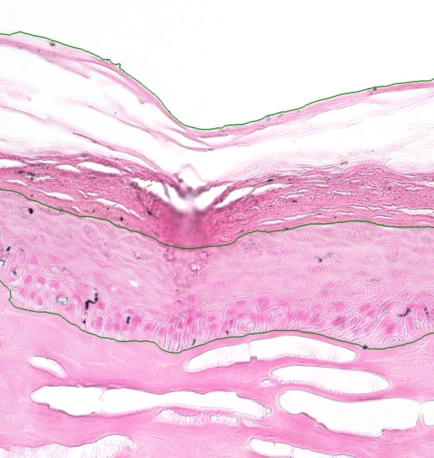}}
%  \centerline{(a) Image example with unbroken}\medskip
\end{minipage}
\begin{minipage}[b]{0.48\linewidth}
  \centering
\centerline{\includegraphics[width=5cm]{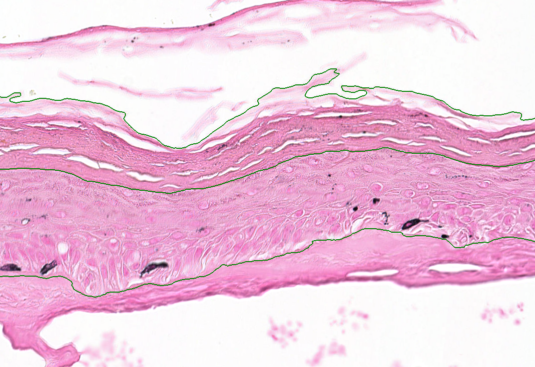}}
   %% \centerline{(a) Result 1}\medskip
\end{minipage}
\begin{minipage}[b]{0.48\linewidth}
  \centering
\centerline{\includegraphics[width=5cm]{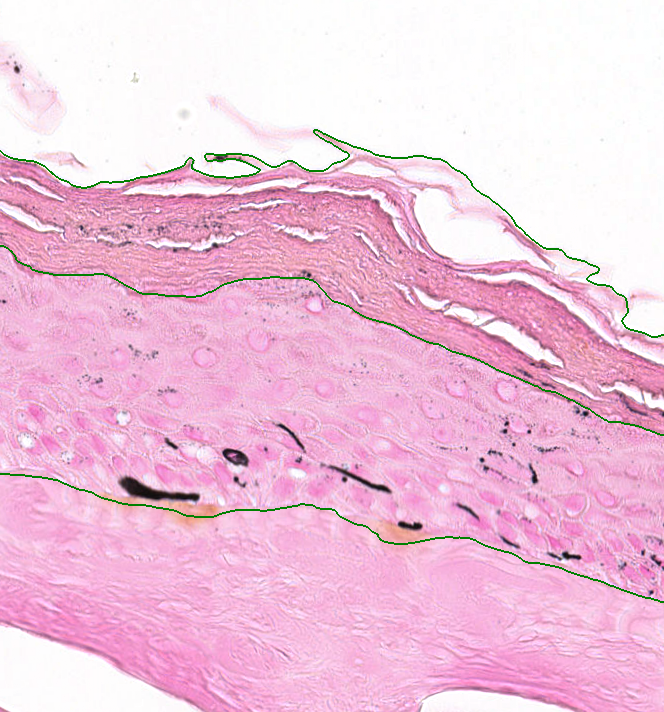}}
%  \centerline{(a) Image example with unbroken}\medskip
\end{minipage}
\begin{minipage}[b]{0.48\linewidth}
  \centering
\centerline{\includegraphics[width=5cm]{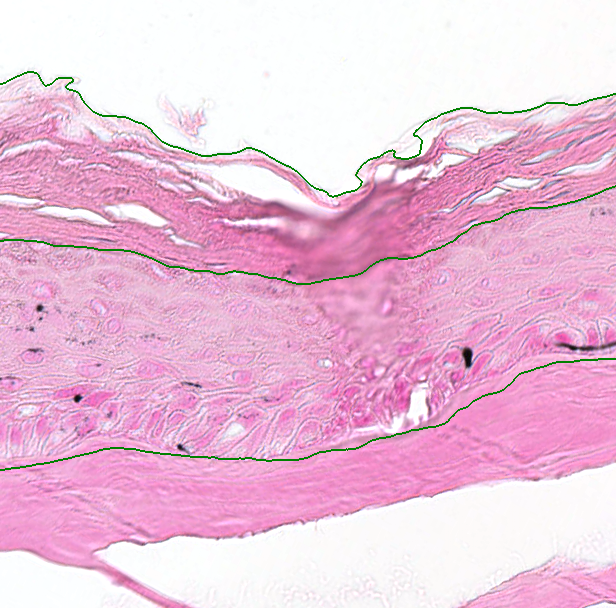}}
   %% \centerline{(a) Result 1}\medskip
\end{minipage}
\caption{Zoom-in on some test images to illustrate the results, as well as its robustness to acquisition artifacts. The contours of the segmentation computed with the final model are overlayed on the original images.}
\label{fig:good}
\end{figure}

It is interesting to note that once a convolutional neural network has been trained (with crops of constant size, as previously stated) it can be applied to images of almost arbitrary sizes.  There are only two limitations: the system memory has to be large enough and neural network architectures that use downsampling layers impose that the dimensions be multiple of some $2n$ (where $n$ is the number of such layers, supposing that the sampling steps are equal to $2$). This approach is interesting not only for practical reasons (no need to compute any more crops and stitch them back together at prediction time) but also significantly alleviates border effects .

\begin{figure}[htb]

  \begin{minipage}[b]{0.48\linewidth}
  \centering
\centerline{\includegraphics[width=5cm]{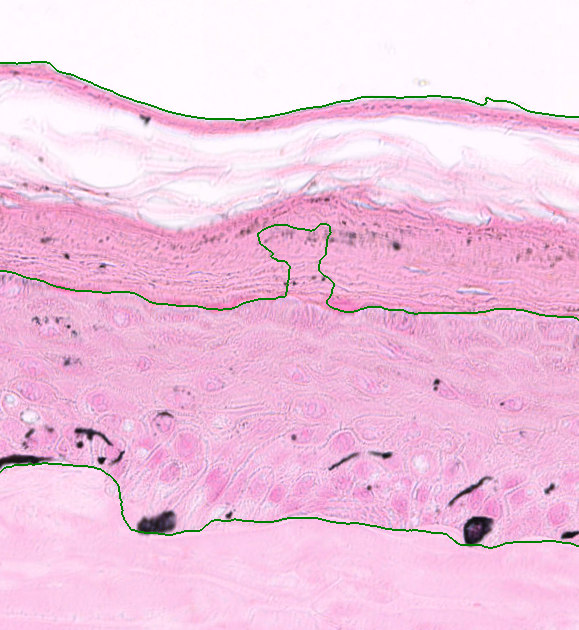}}
%  \centerline{(a) Image example with unbroken}\medskip
\end{minipage}
\begin{minipage}[b]{0.48\linewidth}
  \centering
\centerline{\includegraphics[width=5cm]{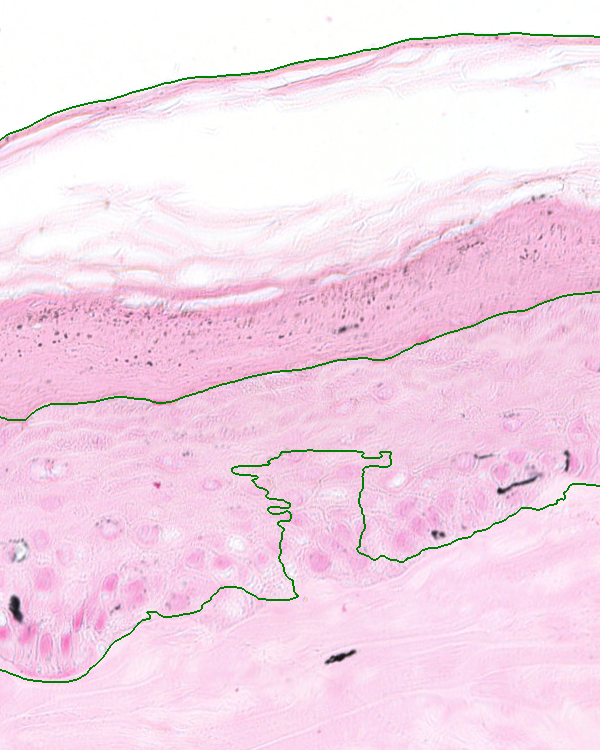}}
   %% \centerline{(a) Result 1}\medskip
\end{minipage}
\caption{Zoom-in onto the two more significant errors found on the 23 images of the test database.}
\label{fig:errors}
\end{figure}

There are $23$ images in the test database. Globally, the results were considered as very good by the final users. They are illustrated in Fig.~\ref{fig:good}. Only two errors were visible at first sight among the $23$ images. They are shown in Fig.~\ref{fig:errors}. Other errors are less visible. They correspond most of the time to a slight displacement of the obtained contour \footnote{One image in Fig.~\ref{fig:good} contains such an error; let the reader try to find it!}.

Tab.~\ref{tab_res} gives some quantitative results on the test database. Accuracy values (the proportion of pixels that are correctly classified) show that incorrectly classified pixels are three times less numerous with our proposed method than with the standard approach. The Jaccard index\footnote{The Jaccard index of two sets is the ratio between the size of their intersection and the size of their union} of the living epidermis region shows almost no improvement: this is natural, as this region can be correctly segmented based solely on local information. On the contrary, the Jaccard index of the \textit{stratum corneum} shows a significant improvement, as the definition of this region heavily relies on non-local information. Finally the mean distance between predicted contours and ground truth (GT) contours confirms this improvement.

\begin{table}[htp]
\centering
\caption{Quantitative results on test database.}
\label{my-label}
\begin{tabular}{|l|c|c|c|c|}
\hline
                & Accuracy & \multicolumn{1}{c|}{\begin{tabular}[c]{@{}c@{}}Jaccard of \\ \textit{stratum corneum}\end{tabular}} & \multicolumn{1}{c|}{\begin{tabular}[c]{@{}c@{}}Jaccard of \\ living epidermis\end{tabular}} & \multicolumn{1}{c|}{\begin{tabular}[c]{@{}c@{}}Mean distance to \\ GT contour\end{tabular}} \\ \hline
Standard U-Net  & 98.33\%                       & 91.46\%                                                                                       & 94.24\%                                                                                        & 18 pixels                                                                                      \\ \hline
Proposed method & 99.49\%                       & 97.43\%                                                                                       & 94.82\%                                                                                        & 4 pixels                                                                                       \\ \hline
\end{tabular}
\label{tab_res}
\end{table}

Processing times are as follows. The standard U-Net takes 171 seconds to process the full 23 test images on a conventional laptop with a NVidia GeForce GTX 980M graphics card. The improved method, including the geodesic reconstruction, takes 407 seconds. We think that the pre-processing could be optimized, but the current version is already fast enough for the application at hand.

\section{Conclusion}

A novel method to utilize global information within a convolutional neural network has been introduced. Based on the morphological reconstruction by dilation, it allows the network to take advantage of geodesic information.

This method has been successfully applied to the segmentation of histological images of reconstructed skin using a U-Net architecture. We believe that a similar improvement should be obtained with other fully convolutional neural networks, such as SegNet.

The method is being integrated in a complete software in order to use it in routine practice.

As a perspective, it would be interesting to explore other ways to use global information within convolutional neural networks, and compare them.

\bibliographystyle{splncs03}
\bibliography{edf_zotero}

\end{document}